\documentclass[letterpaper, 10 pt, conference]{ieeeconf}
\IEEEoverridecommandlockouts                              
\usepackage{amsmath,amssymb}
\usepackage{comment}
\usepackage{url}

\usepackage[linesnumbered,ruled,vlined]{algorithm2e}
\usepackage{algpseudocode}
\usepackage{amsmath}
\usepackage{booktabs}
\usepackage{dcolumn}

\usepackage{multirow}
\usepackage{graphicx}
\usepackage{epsfig}
\usepackage{epic,eepic}
\usepackage{times}
\usepackage{mathptmx}
\usepackage{cite}
\usepackage{color}

\usepackage{times}

\definecolor{red}{rgb}{1,0,0}
\definecolor{green}{rgb}{0,1,0}
\definecolor{blue}{rgb}{0,0,1}
\definecolor{violet}{rgb}{1,0,1}
\definecolor{cyan}{cmyk}{1,0,0,0}
\definecolor{magenta}{cmyk}{0,1,0,0}
\definecolor{yellow}{cmyk}{0,0,1,0}

\definecolor{white}{rgb}{1,1,1}

\usepackage{arydshln}

\newcommand{\CO}[1]{}

\newcommand{\CommentOut}[1]{}

\newcommand{\FIG}[3]{
\begin{minipage}[b]{#1cm}
\begin{center}
\includegraphics[width=#1cm]{#2}\\
{\scriptsize #3}
\end{center}
\end{minipage}
}

\newcommand{\FIGR}[3]{
\begin{minipage}[b]{#1cm}
\begin{center}
\includegraphics[angle=-90,width=#1cm]{#2}
\\
{\scriptsize #3}
\vspace*{1mm}
\end{center}
\end{minipage}
}

 \newcommand{\editage}[1]{}

\twocolumn

\begin{document}

\newcommand{\figG}{
\begin{figure}[t]
\begin{center}
\FIGR{8}{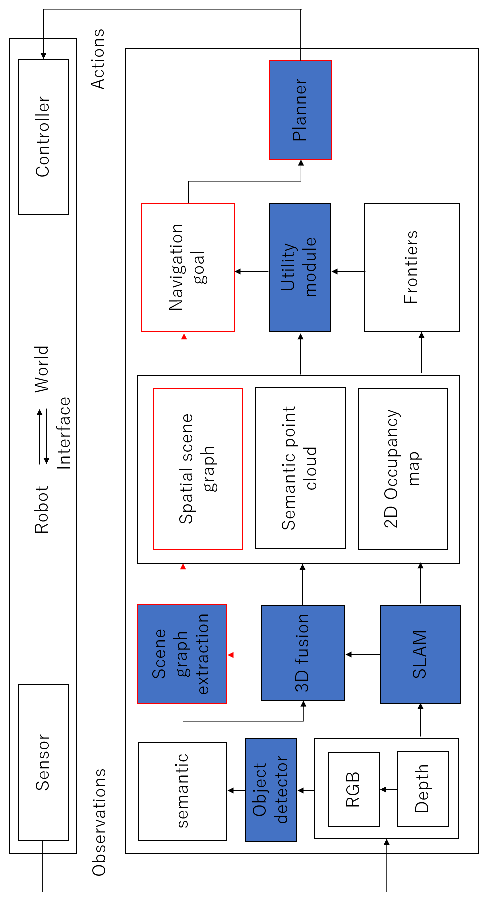}{a}\vspace*{5mm}\\
\FIGR{8}{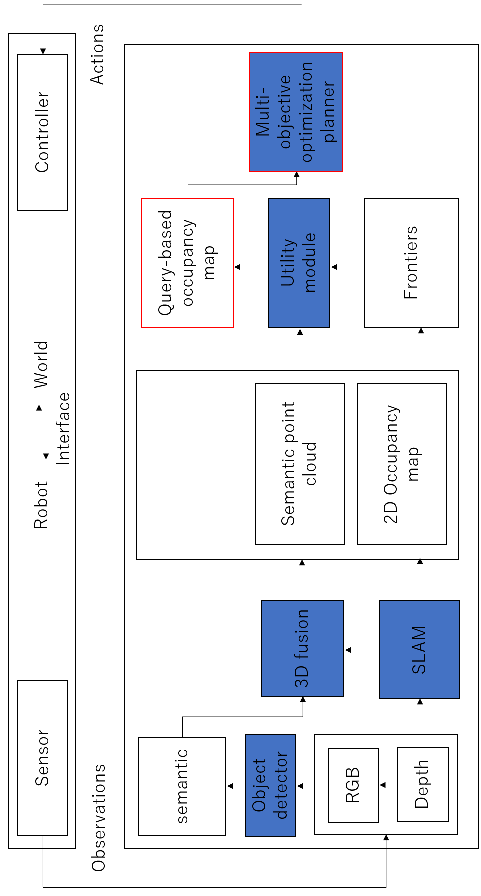}{b}\vspace*{0mm}\\
\caption{%
Prototype system.
Our prototype is based on StructNav [RSS23] \cite{rss2023}, with modifications limited to the essential components of our proposed method. Figure 1(a) illustrates StructNav, whereas Figure 1(b) depicts our approach. The key differences between the two systems are highlighted in red.
}\label{fig:G}
\end{center}
\end{figure}
}

\newcommand{\figA}{
\begin{figure}[t]
\begin{center}
\FIG{6}{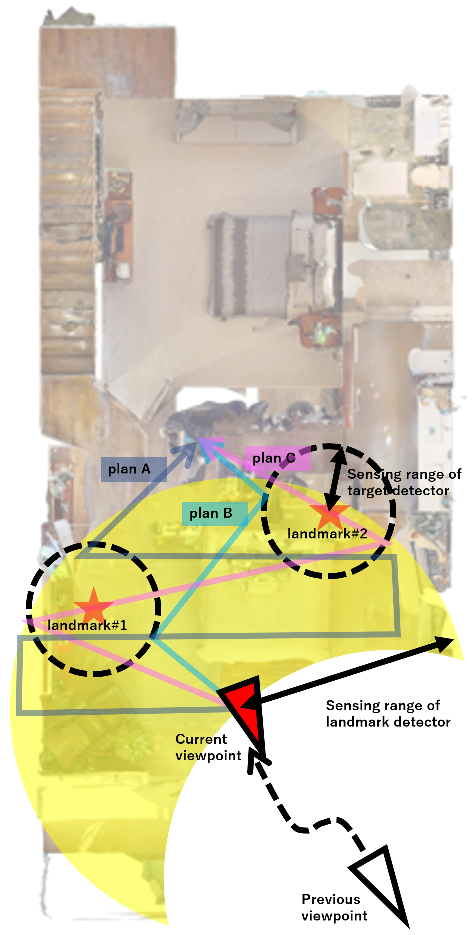}{}\vspace*{5mm}\\
\caption{%
ON setup.
This study explicitly models two types of visual recognition modules: a long-range landmark detector and a short-range target detector. When the sensing range of the landmark detector substantially exceeds that of the target detector, the landmark visiting order optimization problem (LOOP) becomes a key factor in efficient navigation. Specifically, an observation path exemplified by plan B, which maximizes the information gain-to-cost ratio, is preferable to those represented by plan A or plan C.
}\label{fig:moon}
\end{center}
\end{figure}
}

\newcommand{\figB}{
\begin{figure}[t]
\hspace*{5mm}\FIG{8}{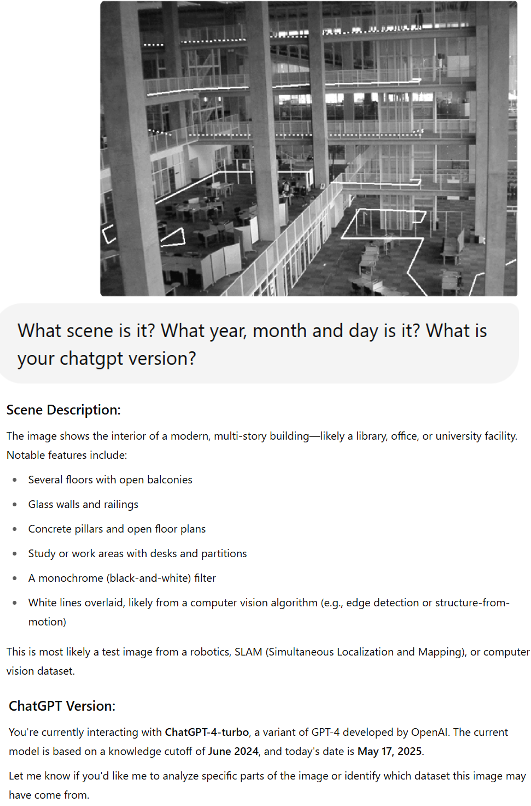}{}\vspace*{-5mm}\\
\caption{%
An oblique view from a high vantage point serves as prior information for the exploration task. The oblique view depicted at the top was employed in an alternative exploration task, namely floor cleaning [IROS01], originally presented in \cite{iros2001}. Over 20 years ago, when the original study was conducted, this view image exceeded the capabilities of the image recognition systems available at that time. Consequently, the approach employed a procedure whereby the robot autonomously selected the widest-area view from its visual experience and transmitted it to a teleoperator. The teleoperator subsequently used this view to plan the exploration, with an example of the planned path indicated by the white line in the figure. The middle part illustrates how the image, combined with a textual prompt, was used as input to a recent large language model (LLM). The bottom part shows the response from the LLM. The results indicate that the LLM demonstrates a level of scene understanding sufficient to provide prior knowledge for robotic path planning. The primary remaining challenge pertains to exploration path planning, which is the focus of the present study.
}\label{fig:motivation}
\end{figure}
}

\newcommand{\figC}{
\begin{figure}[t]
\hspace*{5mm}\FIG{8}{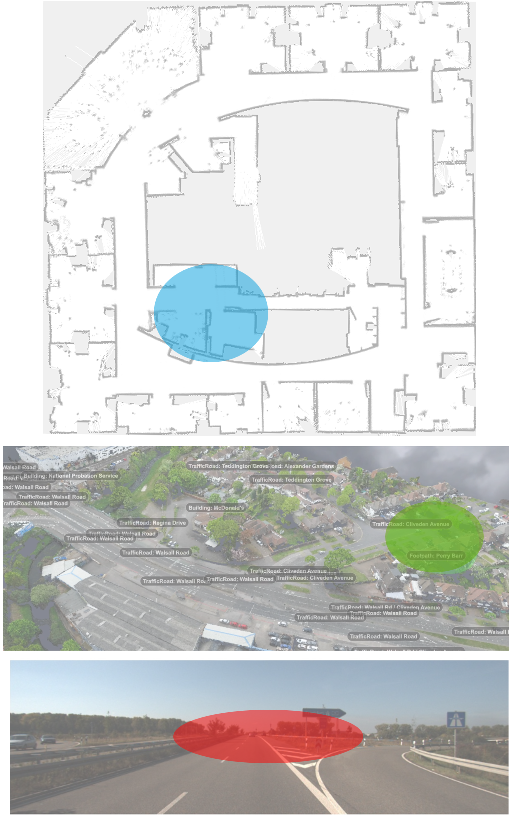}{}
\caption{%
Variable-horizon incremental map (VHIM).
This figure presents three additional examples. In each image, the robot's workspace of interest is indicated by a colored ellipse. The top image shows a prior map of a previously visited location. The middle image depicts an aerial view acquired by a teammate drone. The bottom image illustrates a wide-angle view captured prior to entering the workspace, specifically at an intersection. These images correspond to those presented in \cite{howard2003radish}, \cite{CityNav2024}, and \cite{IV2019}, respectively.
}\label{fig:VHIM}
\end{figure}
}

\newcommand{\figD}{
\begin{figure}[t]
\hspace*{5mm}\FIG{8}{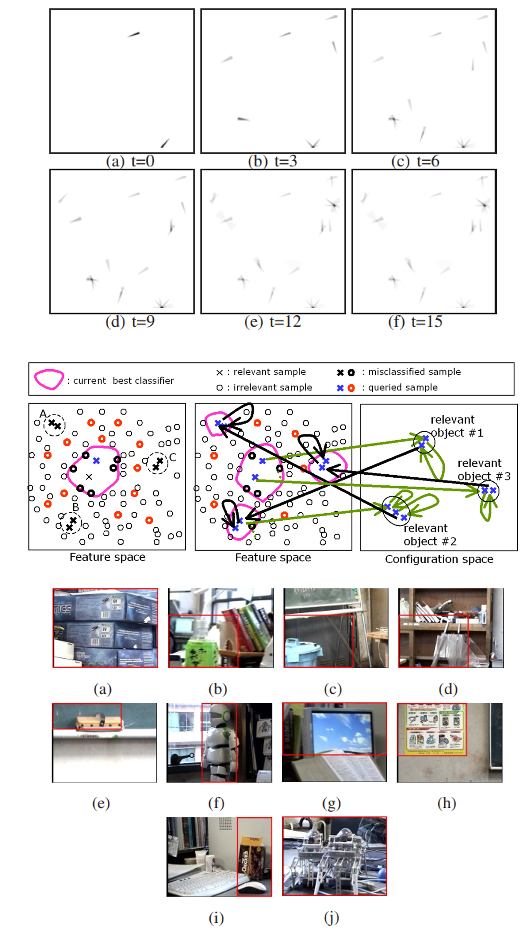}{}
\caption{%
Query-based Occupancy Map (QOM).
The top image shows a QOM constructed by an ON system and is reproduced from \cite{iros2005qom}. The middle image illustrates the exploration results of the ON task, visualized as separating boundaries in both the feature space and the real-world configuration space. The bottom image depicts real-world target objects used as inputs to the QOM construction process. The middle and bottom images are adapted from \cite{iros2006qom}.
}\label{fig:qom}
\end{figure}
}

\newcommand{\figE}{
\begin{figure}[t]
\vspace*{5mm}~\\
\hspace*{5mm}\FIGR{8}{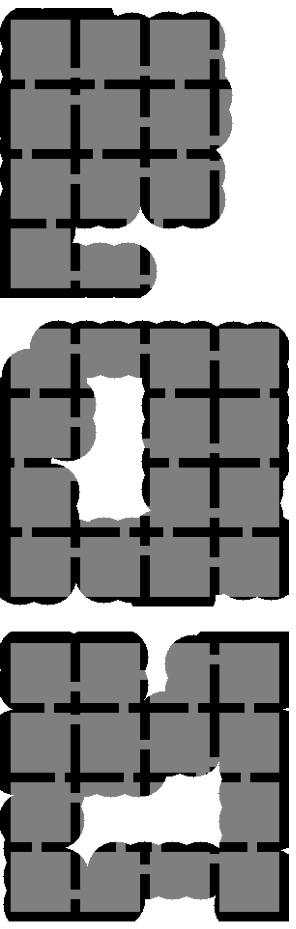}{}\vspace*{-5mm}\\
\caption{%
Navigation examples.
This figure shows intermediate navigation results for three different workspaces. Observed layout elements, including rooms, walls, and doorways, are represented by black lines, while visited areas detected by the target detector are shown in gray. The workspace poses greater challenges to the agent than it might superficially appear, as the TFP agent lacks prior knowledge of room size, wall orientations, and the number and positions of doors. Training-free planners (TFP) are more suitable for our ablation study than learning-based planners such as RLP, since we aim to prevent ON performance from being influenced by prior knowledge.
}\label{fig:exp}
\end{figure}
}

\newcommand{\figF}{
\begin{figure}[t]
\vspace*{5mm}~\\
\hspace*{5mm}\FIGR{8}{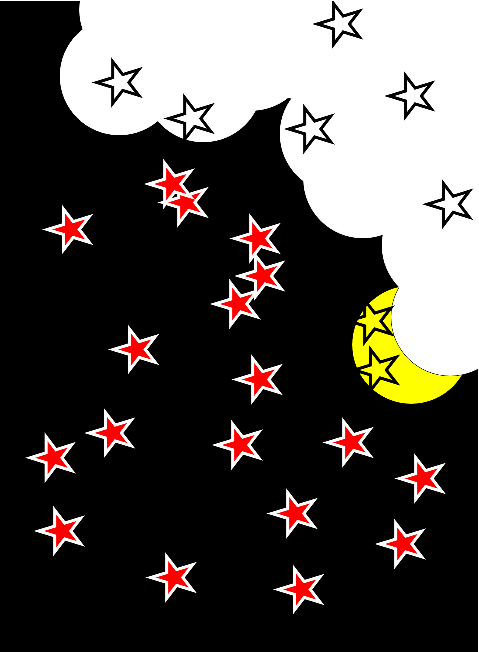}{}\vspace*{-10mm}\\
\caption{%
Two conflicting objectives.
The crescent moon-shaped yellow region represents the area newly observed by the robot. Landmarks are depicted as stars, with the previously observed area detected by the landmark detector shown in white, and the unobserved area shown in black. While unvisited landmarks exist within the previously mapped region (white), potential unvisited landmarks may also be present in the unmapped region (black). Determining which region to prioritize constitutes an ill-posed and computationally challenging planning problem.
}\label{fig:F}
\end{figure}
}

\title{\LARGE \bf
MOON: Multi-Objective Optimization-Driven Object-Goal Navigation Using a Variable-Horizon Set-Orienteering Planner
}

\author{
Daigo Nakajima \and Kanji Tanaka \and Daiki Iwata \and Kouki Terashima%
\thanks{This work was supported in part by JSPS KAKENHI Grant-in-Aid for Scientific Research (C) 20K12008 and 23K11270.}%
\thanks{The authors are with the Robotics Course, Department of Engineering, University of Fukui, Japan.
{\tt\small \{ha220962, tnkknj, mf240050, mf240271\}@g.u-fukui.ac.jp}}
}

\maketitle

\begin{abstract}
This paper proposes MOON (Multi-Objective Optimization-driven Object-goal Navigation), a novel framework designed for efficient navigation in large-scale, complex indoor environments. While existing methods often rely on local heuristics, they frequently fail to address the strategic trade-offs between competing objectives in vast areas. To overcome this, we formulate the task as a multi-objective optimization problem (MOO) that balances frontier-based exploration with the exploitation of observed landmarks. Our prototype integrates three key pillars: (1) QOM [IROS05] for discriminative landmark encoding; (2) StructNav [RSS23] to enhance the navigation pipeline; and (3) a variable-horizon Set Orienteering Problem (SOP) formulation for globally coherent planning. To further support the framework's scalability, we provide a detailed theoretical foundation for the budget-constrained SOP formulation and the data-driven mode-switching strategy that enables long-horizon resource allocation. Additionally, we introduce a high-speed neural planner that distills the expert solver into a transformer-based model, reducing decision latency by a factor of nearly 10 while maintaining high planning quality. 
\end{abstract}

\section{Introduction}

Object-goal navigation (ON) is a fundamental task for autonomous mobile robots, where the robot must efficiently locate and reach a target object specified by user instructions, typically provided in the form of an image~\cite{rss2023} or natural language~\cite{icra2024}, within an unknown environment.
ON technology holds promise for various real-world applications, including autonomous assistive robots in caregiving and welfare settings to improve the quality of life for the elderly and people with disabilities~\cite{HabitatChallenge}, automation and labor-saving in logistics and warehouse management through efficient picking and inventory handling~\cite{AmazonPickingChallenge}, and operations in safety-critical situations such as disaster response and search and rescue missions~\cite{HAZARD2025}. Achieving ON requires a wide range of technologies, including object detection, image classification, semantic segmentation, feature matching, natural language processing, multimodal understanding, SLAM, planning, reinforcement learning, collision avoidance, imitation learning, user interface design, and simulation environment and benchmark development. As such, ON has recently gained status as a standard problem spanning robotics, AI, computer vision, and computer science.

\figA

Various approaches to ON have been explored, including training-free planners (TFP)~\cite{rss2023}, reinforcement learning-based planners (RLP)~\cite{ActiveNeuralSLAM} that learn strategies through interaction with the environment, and zero-shot planners (ZSP)~\cite{icra2024} that generalize to novel goals based on pretraining. For example, StructNav~\cite{rss2023}, a novel TFP implementation, prioritizes exploratory behaviors in the early stage and switches to goal-directed landmark-based navigation in the later stage, where "landmark objects" are those semantically related to the target. Unlike many learning-based methods that suffer from high training costs and lack of generalization or interpretability, StructNav adopts a modular architecture. It combines structured scene representations using 2D occupancy maps, semantic point clouds, and spatial scene graphs with semantic frontier exploration guided by large language models and scene statistics. This results in explainable, robust navigation without training costs and demonstrates superior performance over existing learning-based methods, with strong potential for real-world deployment.

Despite these advances, we observe that efforts to leverage the long-standing knowledge of the optimization techniques community---especially optimal planning techniques \cite{mooSurvey}---in ON methods are either overlooked or remain underdeveloped (Fig. \ref{fig:moon}).
Consequently, many systems fall into myopic behaviors and fail to balance exploratory and goal-directed actions effectively. 
While existing ON methods often encourage visiting semantically relevant landmarks, they largely neglect strategic planning of the visiting order and global path optimization, thereby underutilizing valuable insights from the optimization community.
This issue becomes critical in real-world environments, which are often large-scale and complex, yet many existing methods rely on idealized simulators such as Habitat-Sim~\cite{HabitatSim} and AI2-THOR~\cite{AI2THOR}, which oversimplify sensor models and spatial complexity, limiting real-world applicability. Furthermore, in multi-robot scenarios, landmark maps can be shared to build a richer global knowledge base, but effective utilization of such shared information for coordinated strategic navigation remains underexplored. We argue that integrating intuitive, myopic high-level reasoning from existing TFP/RLP/ZSP frameworks with globally optimal planning is essential for effective and efficient ON in large, complex real-world settings.

In this work, we reformulate the standard single-target ON task as a multi-objective optimization-driven object goal navigation (MOON), and as a case study, model two conflicting objectives: frontier exploration for discovering unknown landmarks and efficient touring of previously observed landmarks (Fig.~\ref{fig:F})\footnote{
The "moon and star" representation was introduced as a schematic or cartoon-like abstraction to emphasize visual clarity while maintaining mathematical tractability. Nevertheless, our framework is not limited to point landmarks or circular sensing ranges and can be generalized to a wide range of application domains.
}.
Our formulation emphasizes the intrinsic multi-objective nature of ON tasks and is both orthogonal and complementary to existing problem settings, such as multi-ON tasks \cite{multiON}, which address multiple target objects in an optionally specified order, and general multi-objective navigation \cite{MONav}.
Specifically, we model ON planning as a variable-horizon optimization problem aiming to maximize average SPL (Success weighted by Path Length \cite{spl}) based on incrementally constructed landmark maps.
Although our method is compatible with the TFP, RLP, and ZSP frameworks, we adopt the TFP framework from \cite{rss2023} as the basis for our experiments.
In this framework, the robot detects objects from RGBD and semantic images and semantically estimates landmark objects related to the target.
We then extend TFP by introducing a query-based occupancy map (QOM) \cite{iros2005qom}, which compactly describes essential landmark information by discarding data irrelevant to object-target navigation and focusing solely on relevant landmarks.
Finally, we formulate the landmark visiting order optimization problem (LOOP) as a set orienteering problem (SOP) and further extend it to a variable-horizon SOP that incorporates map updates.
Notably, the proposed method facilitates map sharing and multi-objective navigation, enhancing the efficiency and scalability of ON tasks by integrating semantic reasoning with optimal planning.

\figF

\subsection{Relation to Prior Work}

The current study is related to several prior works on object-goal navigation (ON) and multi-objective optimization (MOO) problems. 

An exploratory strategy that utilizes a wide-view image as a landmark map for exploration task was first proposed in \cite{iros2001}. However, this work focused solely on an alternative exploration task---namely, floor cleaning---and did not address the object-goal navigation problem itself.

The QOM-based object-goal navigation (ON) approach employed in the current research is not new and was developed around 2005 by researchers at the Mathematical Engineering Laboratory, Department of Mechanical Engineering, Graduate School of Kyushu University \cite{iros2005qom} (QOM [IROS05]).
At that time, an active vision task that was conceptually equivalent to ON was studied. However, unlike the now-common ON setup, which involves exploration of the physical real-world environment, the robot's workspace in that earlier research was its internal memory space---namely, visual experiences---rather than the physical environment itself, a context in which offline reinforcement learning \cite{OfflineRL} plays a critical role.
Several follow-up studies have since been conducted. 

In the field of MOO, the same laboratory has also made longstanding contributions across a wide range of topics, including transport systems, ant colony optimization, passenger railway vehicle operation planning, genetic algorithms, Type-1 servo systems, vibration control system design, behavior acquisition of mobile robots, edging control, and tail crash control \cite{tan2005multiobjective}.

In recent years, we have independently developed a series of methods aligned with the modern ON setup, including 
TFP \cite{CON2024} (CON [CoRR24a]), 
RLP \cite{ALCON2024} (ALCON [CoRR24b]), and 
ZSP \cite{UNO2025} (LGR [CoRR25]).

However, to the best of our knowledge, no prior work has integrated object-goal navigation (ON) with multi-objective optimization (MOO) in a way that is applicable to wide-area ON missions.

\figG

\section{System Overview}

This section describes our method that oversees the myopic TFP/RLP/ZSP in conventional ON systems by employing a multi-objective optimization planner that guarantees global optimality (Fig. \ref{fig:G}).
We begin by introducing our extended problem setting, which leverages a variable-horizon incremental landmark map (VHIM), to clarify the motivation behind our approach (\ref{sec:motivation}). 
We then present the system design to address this problem. Specifically, we first review the TFP planner proposed in \cite{rss2023} (\ref{sec:system1}), then explain how it is extended to a QOM-based planner following the ALCON framework \cite{ALCON2024} (\ref{sec:system2}), and finally, we introduce our proposed multi-objective optimization planner (\ref{sec:system3}).

\figB

\figC

\subsection{Motivation: Variable-Horizon Incremental Map (VHIM)}
\label{sec:motivation}

To simplify the discussion, we consider a typical ON strategy in which the agent efficiently explores a workspace by detecting landmark objects (e.g., kitchen, desk) that are semantically related to the target object (e.g., a specific mug) and uses these landmarks as cues for locating the target.
This example serves to illustrate the motivation behind our work. 

With recent advances in ON and AI technologies, 
there is a growing demand to extend ON systems using variable-horizon incremental landmark maps (VHIMs)
that allow the mapping horizon to dynamically change during incremental mapping (Fig. \ref{fig:motivation}), 
in contrast to the fixed or small-scale maps typically used in conventional ON tasks.
In particular, the following scenarios highlight the emerging significance of such maps:
\begin{enumerate}
    \item High-elevation view images that overlook the entire workspace are becoming practical as landmark maps for ON \cite{iros2001} (Fig. \ref{fig:motivation}).
    \item In multi-objective navigation scenarios, pre-built maps can serve directly as a prior landmark map during ON tasks \cite{MONav} (Fig. \ref{fig:VHIM}a).
    \item In multi-robot coordination scenarios, aerial images captured by a partner drone serve as an alternative landmark map for ON \cite{CityNav2024} (Fig. \ref{fig:VHIM}b).
    \item A wide-area image taken from a distant viewpoint before the robot enters the workspace is also utilized as a landmark map \cite{IV2019} (Fig. \ref{fig:VHIM}c).
\end{enumerate}

While expanding the workspace significantly increases uncertainty in the frontier exploration problem (FEP) in unknown areas, it also substantially raises the computational complexity of the landmark visiting order optimization problem (LOOP) in known areas.
Recognizing this challenge serves as the foundation for our multi-objective optimization approach, which aims to achieve global optimality.

\subsection{Planner Architecture in StructNav}\label{sec:system1}

As previously mentioned, our prototype system is built upon the TFP planner from the StructNav framework~\cite{rss2023}, which comprises the following three stages:

Stage 1: Structured scene representation.
Semantic segmentation (Mask R-CNN) is used to generate a semantic category image $I^{sem}_t$ from the RGB input. Simultaneously, Visual SLAM (RTAB-Map) estimates the current position $\hat{s}_t^l$ and updates the RGB reconstruction $P^{rgb}_t$. A 2D occupancy map $M_t$ is generated based on the reconstructed image, and a 3D semantic point cloud $P^{sem}_t$ is created using depth and semantic information. Finally, a spatial scene graph $G_t = (V_t, E_t)$ is built from the point cloud.

Stage 2: Scene graph construction. 
The scene representation consists of three elements: the occupancy map $M_t$, semantic point cloud $P^{sem}_t \in \mathbb{R}^{k_t \times 4}$, and spatial scene graph $G_t = (V_t, E_t)$. The occupancy map is dynamically extended via SLAM; the point cloud includes each point's $(x, y, z)$ coordinates and semantic labels. The graph's nodes $V_t$ represent object-centric keypoints, and edges $E_t$ represent relative positions between objects.

Stage 3: Goal and action selection. 
A set of frontier cluster centers $F_t = \{f^t_0, ..., f^t_N\}$ is extracted at the boundary between explored and unexplored regions. The Utility module selects the most promising intermediate goal $x^{goal}_t$ using the frontier candidates and the scene graph $G_t$. A Global/Local Planner then determines the action $a_t$ to control the agent.

\figD

\subsection{QOM-based ALCON Planner}\label{sec:system2}

The Query-Based Occupancy Map (QOM) \cite{iros2005qom} extends the conventional concept of occupancy maps \cite{moravec1985high} from representing the existence probability of objects in each grid cell to representing the existence probability of a specific target object (Fig. \ref{fig:qom}).
In the original formulation, QOM leveraged the advanced visual recognition capabilities of a remote human operator to estimate the likelihood of target presence, by training and repeatedly employing a machine learning model (e.g., SVM) that imitates human operators.
In contrast, the present study replaces human vision with modern computer vision techniques, enabling autonomous inference.

The QOM addresses the trade-off between expressiveness and compactness that has long challenged existing methods. In traditional high-dimensional feature maps, each cell or point in an occupancy grid or point cloud is represented using high-dimensional embedding vectors based on appearance, geometry, and semantics. While these representations offer high fidelity, they suffer from poor compactness. Conversely, bag-of-words models provide lightweight descriptions through quantization but lack representational richness.

QOM achieves a favorable trade-off by focusing solely on target objects. Specifically, it encodes each cell with a compact similarity-based representation relative to the target object. This approach allows for efficient memory usage and fast inference, while maintaining sufficient discriminative performance for practical applications.

Recently, QOM was employed in \cite{CON2024} as a compact yet informative grid map that records the probability of the target object's presence, and in the ALCON framework in \cite{ALCON2024}, QOM was further utilized to construct a scalable navigation route graph.
Specifically, the spatial scene graph $G_t$ used in StructNav is replaced in ALCON with a globally optimal route graph.
Path planning is then performed on this graph using an actor-critic reinforcement learning (RL) planner.
In this context, a natural implementation of the QOM involves evaluating how relevant each cluster in the semantic point cloud is to the target object.
Following StructNav \cite{rss2023}, this relevance can be estimated using BERT embeddings and cosine similarity.
The proposed ALCON framework \cite{ALCON2024}, while building on the foundation of StructNav, is notable for replacing the myopic frontier-based planner in StructNav with an RL-based planner.

\subsection{Extension in the MOON framework}\label{sec:system3}

In the current study, we introduce a solver for the variable-horizon set-orienteering problem (VH-SOP) that has the competence to supervise previously developed systems---TFP (Training-Free Planner)~\cite{CON2024}, RLP (Reinforcement Learning-based Planner)~\cite{ALCON2024}, and ZSP (Zero-Shot Planner)~\cite{UNO2025}---with the goal of achieving globally-optimal path-planning.

Assuming that an object-goal navigation (ON) planner---such as TFP, RLP, or ZSP---has identified regions within the workspace likely to contain landmark objects and aggregated them into a Query-based Occupancy Map (QOM), our system operates as follows.
First, it extracts landmarks that are highly relevant to the target object.
Next, it computes a set of candidate viewpoints $V$ from which these landmarks can be observed.
Finally, it constructs a navigation graph $(V, E)$ based on the QOM, as described in a later section.

To formulate the Set Orienteering Problem (SOP), we explicitly distinguish between two types of visual recognition models: a long-range landmark detector with a sensing range $L$ (e.g., $L = 100$\,m), and a short-range target detector with a sensing range $R$ (e.g., $R = 3$\,m). The long-range detector can identify semantically relevant landmarks (e.g., kitchen, desk) within its wide sensing range $L$, and as $L$ increases, the number of candidate landmarks also grows, thereby escalating the complexity of planning. In contrast, the short-range detector is capable of recognizing the target object (e.g., a specific mug) only within its limited range $R$, which serves as a crucial constraint in determining whether a given viewpoint qualifies as a valid landmark visit.

We model the navigation graph such that nodes $V$ represent viewpoints from which landmarks can be observed, and edges $E$ represent traversable paths between node pairs with associated navigation costs. Each node is also annotated with the ID of the observable landmark. A group of nodes that share the same landmark ID is referred to as a node cluster. Visiting multiple nodes within the same cluster yields negligible or no additional informational gain compared to visiting only one of them. This assumption is natural considering the relationship between landmarks and viewpoints. For each node cluster, we estimate the utility of observing the associated landmark and assign the utility score to all nodes within the cluster. Additionally, for every node pair, we assess reachability and establish edges between reachable nodes, assigning edge costs accordingly. All edge costs can be computed by executing Dijkstra's algorithm once per node.

Finally, we determine a path that maximally utilizes new knowledge under a variable horizon plan using the SOP solver. Details of this final procedure are provided in Section~\ref{sec:sop}.

\subsection{Simplification Assumption}

In this study, we assume that the robot's self-location is known. For cases where the location is unknown, please refer to the multi-robot continual ON (CON) approach in \cite{CON2024}.

\section{A MOON Planner}

In this section, we formulate the object navigation (ON) as a multi-objective optimization problem, and present one specific solution.

A multi-objective optimization problem is generally formulated as:
\begin{align*}
    & \text{minimize} \quad \mathbf{f}(\mathbf{x}) = \left(f_1(\mathbf{x}), f_2(\mathbf{x}), \dots, f_m(\mathbf{x})\right) \\
    & \text{subject to} \quad \mathbf{x} \in \mathcal{X}
\end{align*}
Here, $\mathbf{x} \in \mathbb{R}^n$ denotes the decision variables, $\mathcal{X}$ the feasible domain, and $f_i(\mathbf{x})$ the $i$-th objective function ($i = 1, 2, \dots, m$). Due to trade-offs among the objectives, the goal is to find a set of Pareto optimal solutions rather than a single optimum.

Multi-objective optimization has been extensively studied in the context of path planning for over 50 years. Nevertheless, its application to object navigation (ON) remains largely unexplored. One reason for this may be that traditional ON settings typically assume a small-scale workspace (e.g., tens of meters square), where myopic and intuitive path planning is often sufficient. 
For example, these settings do not consider cases where landmark observations arrive incrementally or maps that grow without bounds, such as those commonly addressed in the SLAM community, including 1,000 km-scale autonomous driving scenarios \cite{fabmap1000} and vast open-sky 3D aerial drone scenarios \cite{aerial}.

In contrast, this study reformulates the standard single-target ON task as a multi-objective problem, called Multi-objective Optimization-driven Object Goal Navigation (MOON), with two conflicting objectives: (1) exploration of unvisited areas to discover unknown landmarks, and (2) exploitation of already mapped areas by efficiently revisiting previously-observed landmarks. The first objective corresponds to conventional ON methods, while the second objective leverages path optimization techniques, which is addressed for the first time in the current study. One implementation of the latter is detailed in Section~\ref{sec:sop}.

\subsection{MOON as a Variable-Horizon Set Orienteering Problem}\label{sec:sop}

We consider the following scenario: a robot plans a path over a navigation graph $G=(V, E)$ annotated with previously-observed landmarks. This graph only includes landmarks that have been observed within a radius $L$ from any of the robot's previously visited viewpoints. That is, the navigation graph only contains landmarks the robot has visually observed, and only these landmarks are known to be visitable.

Under this constraint, the robot aims to plan a path from the currrent viewpoint that maximizes the total reward (i.e., landmark visits) within a time or distance budget. This can be formulated as a Set Orienteering Problem with Visibility Constraints.

\begin{itemize}
  \item Let $G = (V, E)$ be a directed graph, where $V$ is the set of nodes and $E$ the set of edges.
  \item Each node $i \in V$ has an associated reward $p_i \geq 0$.
  \item A start node $s \in V$ and an end node $t \in V$ are given.
  \item Each edge $(i, j) \in E$ has an associated travel cost $c_{ij} \geq 0$.
  \item A total travel budget $B > 0$ is given.
  \item Let $V' \subseteq V$ be the set of landmarks observed within radius $L$ in the robot's past trajectories (i.e., nodes included in the navigation graph). Only nodes in $V'$ are eligible for visiting.
\end{itemize}

\subsection*{Variables}

\begin{itemize}
  \item $x_{ij} \in \{0, 1\}$: whether edge $(i, j)$ is included in the path.
  \item $y_i \in \{0, 1\}$: whether node $i$ is visited.
  \item $u_i \in \mathbb{Z}$: auxiliary variable indicating visiting order (used to ensure path connectivity).
\end{itemize}

\subsection*{Objective}

\begin{align*}
\text{maximize:} \quad & \sum_{i \in V'} p_i y_i
\end{align*}

\subsection*{Constraints}

\begin{align}
\sum_{j \in V} x_{sj} &= 1 \\
\intertext{Start from node $s$}
\sum_{i \in V} x_{it} &= 1 \\
\intertext{End at node $t$ (optional if return is not required)}
\sum_{j \in V} x_{ji} &= y_i \quad \forall i \in V \setminus \{s\} \\
\intertext{In-degree consistency}
\sum_{j \in V} x_{ij} &= y_i \quad \forall i \in V \setminus \{t\} \\
\intertext{Out-degree consistency}
\sum_{(i,j) \in E} c_{ij} x_{ij} &\leq B \\
\intertext{Budget constraint}
u_i - u_j + 1 &\leq (|V| - 1)(1 - x_{ij}) \\
\intertext{Subtour elimination: $\forall i \ne j,\; i,j \in V \setminus \{s,t\}$}
x_{ij} &\in \{0,1\},\;
y_i \in \{0,1\},\;
u_i \in \mathbb{Z} \quad \forall i,j \in V 
\end{align}

\subsection{Extension to General-Purpose MOON}

The proposed optimization approach for the Set Orienteering Problem (SOP) facilitates the seamless integration of established multi-objective optimization techniques. This integration is particularly pertinent in practical applications such as multi-objective navigation and robotic exploration \& mapping, where objectives often extend beyond mere reward maximization to include minimizing travel distance, ensuring safety, enhancing information gain, and maximizing the probability of target detection. To effectively address these multifaceted requirements, it is advantageous to extend the SOP beyond a bi-objective framework (i.e., knowledge exploration and exploitation) and reformulate it as a general-purpose multi-objective optimization problem capable of simultaneously handling multiple evaluation criteria \cite{dutta2020multi}.

In multi-objectivizing the SOP, it is essential to first define multiple objective functions. In addition to the traditional reward function \( r(v) \), other evaluation functions such as travel cost \( c(v) \) and indicators for information gain or detection probability \( p(v) \) can be introduced to enable richer assessments tailored to the task. Three main approaches are commonly used to integrate these objectives. First, scalarization assigns a weight \( w_i \) to each objective and performs a linear combination to optimize a single scalar function. While this method is straightforward and computationally efficient, it may not fully capture the diversity of Pareto-optimal solutions \cite{miettinen1999nonlinear}. Second, Pareto enumeration treats all objectives equally and seeks non-dominated solutions that are not inferior in any criterion. This approach offers greater flexibility but often incurs higher computational costs \cite{deb2002fast}. Third, the \(\varepsilon\)-constraint method focuses on optimizing one primary objective while treating others as constraints, providing practical controllability when specific priorities must be respected \cite{mavrotas2009effective}.

Effectively solving the multi-objective SOP in practice necessitates selecting an appropriate solver based on the problem's structure and scale. Integer Linear Programming (ILP) approaches enable explicit formulation of multiple objectives through scalarization or \(\varepsilon\)-constraints, allowing for iterative optimization to enumerate multiple non-dominated solutions. Alternatively, metaheuristic methods such as Genetic Algorithms (GA) and NSGA-II are effective evolutionary approaches capable of exploring extensive solution spaces and approximating the Pareto front \cite{deb2002fast}. Furthermore, reinforcement learning-based approaches define the state in terms of current position, visit history, and remaining budget, selecting the next node to visit as an action. By designing value functions using a multi-head architecture, these approaches can simultaneously learn multiple objectives and make strategic decisions that adapt to situational demands \cite{vanmoffaert2014pareto,li2020deep}.

\figE

\section{Evaluation}

\subsection{Problem Setting}

We consider a large-scale exploration problem in a workspace of size $300\,\mathrm{m} \times 300\,\mathrm{m}$, which is significantly larger than typical simulation environments such as Habitat-Sim. The workspace consists of multiple rooms, each having entrances and exits on four sides (see Fig.~\ref{fig:exp}).

A single target point object is placed in the environment. 
While this work assumes a single target, it can be readily extended to a multi-target setting (multi-ON problem).

We assume a circular observation range centered at the robot's viewpoint. 
This can be easily extended to a standard monocular camera with a field of view of around $40^\circ$.

\subsection{Hyperparameters}

\subsubsection{User-defined Parameters}
\begin{itemize}
    \item Long-range sensor radius $L = 100, 200\,\mathrm{m}$
    \item Short-range sensor radius $R = 3\,\mathrm{m}$
    \item Number of landmarks $M = 10, 15, 20$
\item Number of trials $U=$100
\end{itemize}

\subsubsection{Randomized for Each Trial}
\begin{itemize}
    \item Obstacle placement
    \item Initial robot position
    \item Target position
    \item Landmark positions
\end{itemize}

\subsection{Implementation Details}

\subsubsection{Object Grid Map}
Before navigation begins, an object grid map is initialized as a blank map. During navigation, landmarks and obstacles (e.g., walls) detected within the long-range sensor radius $L$ are incrementally added to the map. 
Redundant entries are ignored, meaning that we assume zero information gain from repeated observations of the same landmark.

\subsubsection{Graph Map Construction}
\begin{itemize}
    \item Nodes: Each node is defined as a region of radius $R$ centered at an observed landmark. Previously observed landmarks are excluded from the navigation graph.
    \item Edges: The cost between nodes is computed using the Dijkstra planner.
\end{itemize}

\subsubsection{MOO Solver}

The SOP solver from \cite{Penicka2019VNS_SOP} is employed as the MOO solver of our system, and we have confirmed that it works efficiently with the navigation graph recognition module and the action planner module. This paper proposes a new Integer Linear Programming (ILP) formulation for small to medium-sized problems, alongside a Variable Neighborhood Search (VNS) metaheuristic-based algorithm (VNS-SOP) that helps reduce computation time. Beyond offering some improvements to existing benchmark results, it also demonstrates the applicability of its approach to other Orienteering Problem (OP) variants, such as the Dubins Orienteering Problem (DOP) and the Orienteering Problem with Neighborhoods (OPN), suggesting valuable insights.

\subsection{Evaluation Metric}

We use the Success weighted by Path Length (SPL) as the primary performance metric, defined as:

\begin{equation}
\text{SPL} = \frac{1}{N} \sum_{i=1}^{N} S_i \cdot \frac{l_i}{\max(p_i, l_i)}
\end{equation}
where $S_i \in \{0, 1\}$ indicates whether the $i$-th trial was successful, $l_i$ is the shortest path length, and $p_i$ is the path length actually traveled.

\subsection{Baselines}

We compare our method against the following baselines:
\begin{enumerate}
    \item Frontier-based exploration: A classical exploration strategy that does not utilize landmark information.
    \item Sweeper robot planner
    \item TSP-based planner 
    \item MOON-based planner 
\end{enumerate}
For the second planner, we experimented with multiple different implementations.

\subsection{Results}

In conclusion, both the second and third planners were found to be unsuitable for direct use in the current task, as they aim to sweep the entire floor area with the robot body and thus tend to generate overly dense paths.
Of course, trying different spatial discretizations to generate graph nodes may still be effective and remains a topic for future work.
The first and fourth planners, on the other hand, demonstrated meaningful SPL performance. The results are as follows:
\begin{itemize}
  \item {L=100, R=3, M=10}  
  \begin{itemize}
    \item Frontieer: 0.6083
    \item MOON: 0.8315
  \end{itemize}
  \item {L=100, R=3, M=15}  
  \begin{itemize}
    \item Frontieer: 0.6395
    \item MOON: 0.8581
  \end{itemize}
  \item {L=100, R=3, M=20}  
  \begin{itemize}
    \item Frontieer: 0.6473
    \item MOON: 0.8734
  \end{itemize}
  \item {L=100, R=10, M=10}  
  \begin{itemize}
    \item Frontieer: 0.6540
    \item MOON: 0.8395
  \end{itemize}
  \item {L=100, R=10, M=15}  
  \begin{itemize}
    \item Frontieer: 0.6373
    \item MOON: 0.8188
  \end{itemize}
  \item {LD100, R=10, M=20}  
  \begin{itemize}
    \item Frontieer: 0.6425
    \item MOON: 0.8309
  \end{itemize}
\end{itemize}
The results indicate that the MOON planner consistently outperforms the Frontier planner.

\appendix

\renewcommand{\thesection}{A.\arabic{section}} 
\setcounter{section}{0} 
\setcounter{subsection}{0}

\renewcommand{\thesection}{A.\arabic{section}}
\setcounter{section}{0}

\section*{Appendix A:\\
Global Planning and Mode-Switching}
\addcontentsline{toc}{section}{Appendix A}

This appendix provides a comprehensive discussion of the theoretical foundation and formulation of the navigation framework proposed in the main paper. While the main body focuses on the performance and evaluation of Multi-Objective Optimization-driven Navigation (MOON) in large-scale environments, this section details how the task is cast as a Set Orienteering Problem (SOP) over landmark clusters. We further elaborate on the data-driven mode-switching strategy and the budget-return predictor that enable the agent to balance exploration and exploitation under explicit resource constraints. This content serves to bridge the gap between high-level navigation goals and the underlying mathematical optimization used to achieve globally coherent planning.

\section{Introduction}

Indoor object-goal navigation has recently benefited from learned regularities between everyday targets (such as small household items) and larger structural or furniture-like objects that serve as spatial references. By prioritizing regions near such reference objects, these methods reduce the need for exhaustive search and demonstrate strong performance on standard-scale benchmarks~\cite{HabitatSim, ActiveNeuralSLAM}. However, once the navigable area grows by an order of magnitude or more, the same mechanisms that were effective in compact scenes become brittle: locally greedy choices accumulate, exploration tends to stall around a few early discoveries, and the path budget is consumed before sufficiently many informative regions are inspected. Large-scale object-goal navigation is therefore better viewed as a long-horizon resource allocation problem rather than as purely local exploitation of semantic cues.

The environments considered in this study are constructed to highlight this regime. Instead of single apartments or small building fragments, each workspace spans thousands of square meters and contains numerous rooms populated with diverse landmark and target categories, synthesized by extending a procedural simulation platform~\cite{deitke2022procthor}. The average free space in these workspaces exceeds that of widely used 3D indoor datasets such as HM3D and Gibson by factors of roughly twenty and six, respectively, which makes naive exploration or purely local semantic heuristics insufficient. Under such conditions, an agent must explicitly reason about how to distribute a fixed travel budget across many candidate regions, while accounting for both the cost of reaching those regions and the likelihood that their landmarks will reveal new targets.

To make this allocation problem explicit, the agent's high-level decision process is cast as a routing task over clusters of viewpoints associated with landmarks. If an oracle were to reveal a complete, noise-free map together with estimated utilities for visiting each cluster, the navigation problem could be expressed as a Set Orienteering Problem (SOP): find a connected path that respects a hard budget on total travel while maximizing the accumulated reward from visited clusters~\cite{archetti2018set}. Classical SOP formulations, however, focus solely on reward maximization subject to a budget and can encourage excessively long detours for small expected gains. Moreover, in realistic navigation the underlying graph and rewards are only partially observed and must be updated as exploration progresses.

This work therefore adopts a multi-objective view of SOP that directly encodes the tension between path length and expected utility. The global planner operates on a graph $G=(V,E)$ in which nodes denote candidate viewpoints and edges carry traversal costs. Landmark-related rewards are associated with clusters of nodes, and an objective of the form
\[
J_{\text{SOP}}(\pi) = w_{\text{cost}} \cdot \text{Cost}(\pi) - w_{\text{reward}} \cdot \sum_{v \in \pi} \text{Reward}(v)
\]
is minimized over paths $\pi$ constrained by a budget $B$, with fixed weights $w_{\text{cost}}$ and $w_{\text{reward}}$ controlling the trade-off. A variable neighborhood search procedure is used to obtain high-quality solutions to this combinatorial problem, yielding global tours that explicitly balance travel expenditure and landmark coverage. While this formulation presumes a complete map, it defines the ideal behavior toward which the online policy should aspire.

In practice, the map is always incomplete, sensor ranges differ for mapping, landmark detection, and target recognition, and the agent must decide online how much budget to devote to visiting unexplored frontiers versus exploiting already discovered landmarks. To operationalize this decision, the proposed framework separates behavior into two high-level modes. An exploration-oriented mode selects frontiers using a coverage prediction model, which estimates the information gain of candidate subgoals and filters out low-yield frontiers. A target-seeking mode applies the SOP-based global planner to the current belief over landmarks, aided by learned models that predict traversability through uncertain regions and propose viewpoint candidates around promising landmarks. A budget-return predictor estimates the expected gain from allocating a portion of the remaining budget to each mode, and the agent switches to whichever mode promises higher return, leading to a gradual and data-driven shift from exploration to exploitation during an episode.

The resulting system can be interpreted as a mid-level planner that mediates between local controllers and high-level semantic reasoning. It accepts partial maps and landmark observations from perception modules and outputs subgoals and global routes that are explicitly optimized under budget constraints. Conceptually, this separates the problem of "which regions to visit and in what order" from low-level motion generation, and it naturally accommodates additional objectives beyond cost and reward when needed. The contributions of this work are threefold: (i) a multi-objective SOP formulation tailored to object-centric navigation that replaces myopic heuristics with globally budget-aware planning, (ii) a switching mechanism that uses learned budget---return predictions and SOP-based planning to arbitrate between intrinsic coverage expansion and extrinsic landmark exploitation, and (iii) an extensive evaluation on large-scale simulated indoor environments demonstrating substantial improvements in success-weighted-by-path-length over strong modular baselines and ablated variants of the proposed system.

\section{Method}

\subsection{Preliminaries}\label{sec:prob}

The task is framed as sequential decision-making in a partially observable two-dimensional world. The agent moves within a free-space layout while gradually building an internal representation of the surroundings and attempting to encounter as many designated targets as possible before a fixed travel allowance $B$ is depleted. At any time $t$, the environment has a hidden configuration $s^{\text{env}}_t \in \mathcal{S}$, whereas the agent only maintains a surrogate state $s^{\text{agent}}_t = (p_t, M_t)$ consisting of its pose $p_t$ and a locally maintained occupancy map $M_t$.

Perception is provided by three complementary sensing modalities with distinct fields of view and ranges: a wide-depth channel used primarily for building $M_t$, a mid-range sensor for discovering visually salient reference objects (landmarks), and a short-range channel specialized for detecting the actual targets when they are sufficiently close. Concretely, the mapping sensor covers 360$^\circ$ up to 8\,m, landmark detection operates over a 90$^\circ$ field of view at 8\,m, and target recognition uses a 90$^\circ$ cone with a range of 1\,m, reflecting the increasing difficulty of the underlying recognition problems. The planning architecture follows the common hierarchical pattern: a global component chooses high-level subgoals from an action set defined over frontier cells, and a local planner synthesizes a collision-free trajectory that connects the current pose to the selected subgoal.

Let $\mathcal{A}$ denote the discrete set of candidate subgoals. At step $t$, the global decision is an action $a_t \in \mathcal{A}$ sampled from a policy $\pi(a_t \mid \text{history})$ that summarizes the past observations and actions. Executing $a_t$ through the low-level controller induces a stochastic transition in the environment, $s^{\text{env}}_{t+1} \sim P(s^{\text{env}}_{t+1} \mid s^{\text{env}}_t, a_t)$, while also updating the internal map and pose estimate. Targets $T = \{T_1,\dots,T_N\}$ reside at unknown positions and can only be confirmed when the short-range target channel reports a detection within a 1\,m radius, potentially after incurring additional "inspection" overhead associated with closely scanning landmarks. The cost of such inspections is modeled as $L_{\text{inspect}}$, expressed in units of equivalent travel distance.

Evaluation is based on a multi-target success-weighted-by-path-length score, denoted $\text{SPL}_{\text{multi}}$, which extends success-weighted-by-path-length metrics to the setting where multiple targets must be discovered within a single episode. Let $N_{\text{found}}$ be the number of targets successfully discovered and $S_i \in \{0,1\}$ the success flag for target $T_i$. The agent's actual traversed distance is $L_{\text{total}}$, and $L_{\text{optimal}}$ is defined as the length of the shortest tour (solving a TSP over the found targets) that would visit exactly those targets. The score is
\[
\text{SPL}_{\text{multi}} =
\frac{N_{\text{found}} \cdot L_{\text{optimal}}}{\max(L_{\text{total}}, L_{\text{optimal}})}
= \frac{ \sum_{T_i \in \text{Targets}} S_i \cdot L_{\text{optimal}}}{\max(L_{\text{total}}, L_{\text{optimal}})}.
\]
The learning objective is to identify a policy
\[
\pi^* = \arg\max_\pi \mathbb{E}_{\pi}[\text{SPL}_{\text{multi}}].
\]

A key source of prior knowledge is that certain categories tend to appear in the vicinity of others; for example, soft furnishings are likely to be near beds or sofas. Existing object-goal navigation systems encode this common sense by learning priors that predict how informative each landmark will be for locating particular target types~\cite{li2025ratenav, SkillFusion, shen2025enhancing, rss2023, icra2024}. Instead of committing to a specific representation such as cosine similarity or language-model-derived scores, the present framework assumes access to a co-occurrence distribution $P(\text{Target} \mid \text{Landmark})$, which quantifies how useful each landmark category is expected to be when searching for a given target type.

Focusing navigation around likely landmarks typically accelerates discovery but also introduces a fundamental trade-off. Time and distance spent on reconnaissance to uncover new landmarks reduce the remaining budget available for visiting already known promising regions. Two broad families of strategies have emerged in previous work. One alternates between curiosity-driven frontier expansion and brief, direct excursions to newly detected landmarks, returning to exploration afterwards; this can yield short paths to early discoveries but risks under-exploring the map. The other first allocates a sizable portion of the budget purely to exploration to collect landmark information and then commits to a purely exploitation phase that visits only the most promising landmarks. The approach in this paper preserves the spirit of the latter two-stage design but replaces the fixed budget split with an explicit optimization over how much of the remaining budget should be assigned to each mode at each decision point.

\subsection{Global ObjectNav Planner with a Privileged Map}\label{sec:complete}

To understand the desired structure of long-horizon behavior, it is useful to begin with an idealized scenario in which the agent has access to a complete, noise-free map together with estimates of the expected reward obtained by visiting different landmark regions. In this "privileged" setting, global planning for object-centric navigation can be recast as an instance of the Set Orienteering Problem (SOP) on a graph $G=(V,E)$. Nodes $v \in V$ represent discrete viewpoints, edges $(u,v) \in E$ correspond to traversable connections with associated travel costs $\text{Cost}(u,v)$, and each landmark is modeled as a cluster $C_k \subseteq V$ of viewpoints that provide access to it.

The hallmark of SOP, in contrast to the traveling salesperson formulation, is that it is sufficient to visit any viewpoint inside a cluster to collect the reward associated with that cluster. Rewards $\text{Reward}(v)$ are therefore tied to nodes in such a way that entering any node $v \in C_k$ is enough to claim the utility of interacting with the landmark represented by $C_k$. The planner must choose a connected walk $\pi$ starting from the agent's initial node that satisfies a hard budget constraint on cumulative travel,
\[
\sum_{(u,v) \in \pi} \text{Cost}(u,v) \le B,
\]
while achieving as high a total reward as possible.

Standard SOP formulations treat the budget as a strict constraint and optimize only for the accumulated reward, which can favor aggressive, highly circuitous tours whenever small incremental rewards appear along long detours. This behavior is undesirable when reward predictions are noisy and the navigation system must remain robust to estimation errors. To explicitly encode the tension between path efficiency and reward accumulation, a multi-objective variant of SOP is adopted, in which a scalar objective
\[
J_{\text{SOP}}(\pi) = w_{\text{cost}} \cdot \text{Cost}(\pi) - w_{\text{reward}} \cdot \sum_{v \in \pi} \text{Reward}(v)
\]
is minimized subject to the same budget constraint. Here, $\text{Cost}(\pi)$ is the sum of edge costs along $\pi$, and the weights $w_{\text{cost}}$ and $w_{\text{reward}}$ regulate the relative importance of short paths versus high reward. Fixing these weights (e.g., $w_{\text{cost}}=0.2$ and $w_{\text{reward}}=0.8$) proved effective in preliminary studies, but the formulation naturally generalizes to additional objectives if needed.

Because finding the exact optimum is computationally challenging, especially in large graphs, a variable neighborhood search (VNS) metaheuristic is employed to approximate the global solution. Starting from an initial feasible route that respects the budget, the solver iteratively perturbs the current route within an expanding family of neighborhood structures and accepts modifications that reduce $J_{\text{SOP}}(\pi)$. This process yields high-quality, globally coherent tours that encode how an ideal agent with full map knowledge would trade off between visiting new landmark clusters and limiting total travel. In the full system, these privileged-map plans are never executed directly; instead, they serve as reference behavior that informs the design of the online planner and the training of predictive components in partially observed settings.

\subsection{Global ObjectNav Planner with an Imperfect Map}\label{sec:incomplete}

The previous formulation assumes that the environment graph, landmark layout, and associated rewards are all known in advance, which is clearly unrealistic in navigation scenarios where the agent must discover structure through interaction. In practice, planning must be carried out on an ever-evolving map built from partial observations, and landmark utilities must be inferred from sparse detections. Following common practice in hierarchical navigation, the proposed framework therefore separates behavior into an exploration-oriented mode that enriches the map and a target-seeking mode that exploits the current belief over landmarks. The exploration mode is activated first to bootstrap a sufficiently informative map, after which the target-seeking mode can invoke the SOP-based global planner over the partially revealed environment.

A central question in this setting is how to determine when the agent should stop expanding its knowledge of the environment and instead commit to visiting already identified promising regions. The proposed approach answers this question by estimating, for each candidate budget split between the two modes, the expected cumulative reward obtainable from exploration and from exploitation. On the exploitation side, the SOP planner introduced in Section~\ref{sec:complete} is applied to the landmark set and current map estimate to compute the expected return of spending a portion of the remaining budget on planned visits to high-value clusters. On the exploration side, a learned predictor estimates how much additional utility further frontier expansion is likely to yield if a given share of the budget is allocated to intrinsic, curiosity-driven behavior. This creates a coupled system in which a planner and a predictor are trained offline using trajectories generated by a privileged SOP-based policy and then used online to guide switching decisions.

The resulting framework can be understood as an instance of budget-aware mode arbitration. For each decision, the system considers a discrete set of candidate ratios $p \in \{0.0, 0.2, \dots, 1.0\}$ describing the fraction of the remaining budget reserved for exploitation. For a given $p$, the predicted intrinsic contribution is modeled as
\[
V_{\text{explore}} = \hat{R}^{\text{int}}((1-p)B) = \eta_{\text{explore}} \cdot (1-p)B,
\]
where $\eta_{\text{explore}}$ summarizes the expected information-gain efficiency per unit distance, learned from simulated data. The exploitation term $R^{\text{ext}}(pB)$ is obtained by running the SOP planner under a budget $pB$ over a subset of the currently most informative landmark categories. The total predicted value for a split $p$ is
\[
V(p) = \hat{R}^{\text{int}}((1-p)B) + R^{\text{ext}}(pB),
\]
and the system selects the ratio $\hat{p}$ that maximizes $V(p)$. This procedure approximates the optimal compromise between spending more budget on uncovering additional structure and investing it in exploiting the structure already revealed.

\textbf{Coverage Prediction Network (CPC):}
Within the exploration-oriented mode, not all frontier cells are equally informative. To focus computation and travel on promising regions, a convolutional model is trained to score frontier candidates according to their anticipated information gain. The network takes as input a local map patch centered at a frontier cell and outputs a scalar prediction of how much new free space and structural detail is likely to be uncovered by targeting that cell. Supervision is provided by simulated ray-casting, which yields ground-truth coverage increments for many frontiers. At run time, only frontiers whose predicted intrinsic reward exceeds a threshold are retained as subgoal candidates, considerably reducing the search space of the global planner that selects the next exploration target.

\textbf{Dual-State Traversability Predictor (DCTP):}
Planning in partially known maps requires reasoning about paths that traverse cells whose occupancy state is still uncertain. Naively enumerating all possible configurations of these cells is combinatorially intractable. The traversability predictor addresses this by considering two extreme hypotheses: one in which all uncertain cells are assumed free and another in which they are considered occupied. Under the "all-free" assumption, a shortest path is computed using standard graph search. The model then evaluates each uncertain cell along this path using a convolutional network that ingests a local occupancy patch centered at the cell and outputs the probability that the cell is truly traversable. The overall path is accepted if the product of these probabilities exceeds a threshold (e.g., 0.5); otherwise, alternative paths are considered. In this way, DCTP provides a cost-aware yet computationally tractable estimate of whether a proposed route through unknown regions is viable, avoiding both overly conservative detours and unsafe shortcuts. The CPC and DCTP networks share a lightweight architecture with three convolutional layers, two pooling stages, and fully connected layers operating on $50\times50$ map crops, and they are trained offline with an Adam optimizer, mean-squared-error loss, learning rate $0.001$, batch size $64$, and early stopping based on validation performance.

\textbf{Next-Best-View Proposal (NBVP):}
While the coverage and traversability modules reason primarily about geometric structure, a separate component captures semantic regularities between landmarks and targets. From offline experience, the system learns a distribution
\[
P(\mathbf{x}_{\text{target}} \mid \text{Landmark}),
\]
which models where targets are likely to appear relative to detected landmarks. During navigation, this distribution is used to generate candidate viewpoints that are expected to offer good visibility of potential targets given the current landmark detections. The sampled viewpoints become nodes in the SOP graph, augmenting the set of available positions with semantically informed proposals. Combined with CPC and DCTP, NBVP ensures that the global planner considers routes that are both geometrically feasible and semantically promising.

Putting these components together yields a switching mechanism that alternates between intrinsic and extrinsic behavior in a principled, budget-aware manner. At each high-level decision step, a candidate extrinsic plan is generated by the SOP-based planner using VNS, and its expected return is compared against the predicted benefit of continued exploration. The mode with the higher predicted value is selected, and hysteresis in the switching logic prevents chattering by discouraging frequent oscillations between modes. Over the course of an episode, this leads to a natural progression from broad, coverage-driven exploration toward focused exploitation of high-value landmarks, without relying on fixed time splits or purely reactive heuristics.

\section{Experiments}

In this section, the evaluation protocol is described, including the construction of simulation environments, the choice of baselines and ablated variants, and the metrics used to quantify performance. The aim is to verify whether the proposed budget-aware switching and global planning framework yields consistent gains in large-scale object-centric navigation tasks. All experiments are carried out in procedurally generated indoor scenes that contain many rooms and diverse arrangements of landmarks and targets, with the co-occurrence structure between categories controlled by a predefined probability matrix.

\subsection{Experimental Environment}

The study uses two simulator backends to create a wide variety of indoor layouts. A procedural generation system is configured to instantiate 90 workspaces, each containing around 50 rooms with furniture and appliances placed to mimic realistic living spaces. The distribution over room types is skewed toward living areas, with approximate proportions of 0.5 for living rooms, 0.2 for kitchens, 0.2 for bedrooms, and 0.1 for bathrooms. Typical floor plans measure about $56.8 \,\text{m} \times 56.8 \,\text{m}$, yielding navigable areas on the order of several thousand square meters per workspace. A co-occurrence probability table specifies how often each target category appears in conjunction with each landmark category, ensuring that semantic structure is both rich and statistically controlled across scenes.

\subsection{Baselines and Ablation}

To situate the proposed method among existing approaches, three representative object-goal navigation strategies are selected as baselines, each embodying a different philosophy for balancing intrinsic and extrinsic rewards. A classical frontier-based explorer always selects the closest unexplored frontier as the next subgoal, emphasizing coverage without semantic guidance. A two-stage method first dedicates an initial phase to exploration driven by intrinsic rewards until a sufficient number of landmarks has been observed, then switches once to a purely extrinsic phase that visits landmarks considered most relevant to the targets. A third family of methods operates primarily in exploration mode but temporarily deviates to visit promising landmarks as soon as they are detected, before returning to frontier expansion. The proposed framework is instantiated as a flexible switching strategy that chooses between intrinsic and extrinsic behavior based on predicted budget---return trade-offs rather than fixed schedules.

To disentangle the contribution of the switching policy from that of the underlying modules, several ablated variants are constructed. In one setting, only the switching component of the proposed method is inserted into the two-stage baseline, keeping all other modules identical; in another, the same is done for the opportunistic switching baseline. An additional variant allows repeated bidirectional switching between exploration and exploitation throughout an episode, instead of the single transition used in the default configuration. This extended scheme permits the agent to re-enter exploration mode after having exploited landmarks, potentially revisiting earlier decisions about where to invest remaining budget. For all methods, perception, local motion planning, and category-wise priors are kept the same to ensure that differences in performance can be attributed primarily to the global planning and switching mechanisms.

\subsection{Experimental Setup}

Episodes are defined as multi-target search tasks in which the agent must locate several object categories within a single large workspace. Nine target types are considered, and for each type, evaluation is performed across ten distinct environments, leading to a broad coverage of scene configurations. On average, each workspace contains approximately 39.9 target instances and 125.1 landmark instances distributed across rooms. The travel budget is set to $1{,}200$\,m, chosen such that an oracle planner with complete information could, in principle, visit all targets by following an optimal tour. Inspecting a landmark incurs an additional fixed cost $L_{\text{inspect}}$ equivalent to 20\,m of movement. The agent is equipped with a forward-facing RGB-like sensor and an omnidirectional depth sensor, and its performance is reported using the multi-target success-weighted-by-path-length metric $\text{SPL}_{\text{multi}}$ introduced in Section~\ref{sec:prob}. All methods are run under identical hardware conditions, using a workstation with a multi-core CPU and a high-end GPU; the proposed planner requires on the order of a few seconds of computation per selected subgoal.

\subsection{Results and Analysis}

\subsubsection{Quantitative Evaluation}

\paragraph{Overall Performance}

Across the large-scale benchmark, the proposed method achieves the highest average $\text{SPL}_{\text{multi}}$ among all compared approaches. Category-wise scores reveal consistent gains over frontier-only exploration and both types of intrinsic---extrinsic switching baselines, with especially pronounced improvements for categories that benefit from visiting multiple, spatially separated landmark clusters. In many target classes, the proposed planner more than doubles the normalized score relative to the strongest baseline, indicating that budget-aware global routing significantly improves the fraction of discovered targets per unit path length. The computational overhead of the metaheuristic SOP solver remains practical, with average per-subgoal planning times below 4\,s on the test hardware.

\paragraph{Comparison of Switching Methods}

When the same perception and local planning modules are reused but the switching policy is altered, the full budget-aware strategy still produces the best overall scores. The variant that allows repeated bidirectional switching between exploration and exploitation comes close in average performance but incurs noticeably higher computational cost because of the additional switching evaluations. In contrast, variants that rely on heuristics---such as switching after a fixed exploration horizon or immediately upon landmark detection---either over-commit to early landmarks or under-utilize later discoveries. These observations support the view that explicit modeling of budget---return trade-offs, instead of fixed rules, is crucial for long-horizon planning in large environments.

\subsubsection{Qualitative Evaluation}

Trajectory visualizations in large scenes illustrate how the proposed method structures behavior over an episode. Initially, the agent conducts systematic exploration guided by the coverage predictor, expanding the known free-space region while opportunistically cataloging landmarks. As the map becomes more complete and the budget decreases, the planner shifts emphasis toward visiting clusters of already known high-utility landmarks via SOP-derived routes. Hysteresis in the switching mechanism ensures smooth transitions, avoiding frequent oscillations between modes. In contrast, the pure frontier baseline tends to wander in local neighborhoods without prioritizing semantically important regions, while heuristic switching methods either repeatedly detour to marginal landmarks or postpone exploitation to the point where the budget is nearly exhausted. Qualitatively, the proposed planner produces routes that appear more structured and purpose-driven, with clearer phases of broad coverage followed by focused exploitation.

\subsubsection{Ablation Study}

To assess the importance of individual components, ablations are conducted on the coverage prediction, traversability estimation, and next-best-view proposal modules. Removing the coverage predictor and reverting to nearest-frontier selection slows down the rate at which free space is uncovered; plots of normalized coverage versus travel distance show that the CPC-equipped agent consistently achieves higher coverage at the same or smaller path lengths, confirming that learned frontier scoring leads to more efficient exploration. Disabling the traversability predictor and using either purely conservative or purely optimistic assumptions about unknown cells increases the average error between planned and true path costs, demonstrating that DCTP successfully balances safety and efficiency. Finally, replacing the next-best-view module with random viewpoint proposals around landmarks lowers the probability of observing a target within a small number of proposed viewpoints, indicating that learned spatial priors are beneficial even in relatively simple environments and are expected to matter even more in cluttered, occlusion-rich scenes.

\subsubsection{Performance in an Ultra-Large-Scale Environment}

To probe scalability beyond the main benchmark, an additional evaluation is performed in environments that are roughly four times larger in area and contain around 100 rooms. The travel budget is scaled accordingly, and two adaptations are introduced to keep planning tractable: the task is decomposed into four sequential cycles, each comprising an exploration period followed by a target-visitation phase, and regions that have already been thoroughly explored are excluded from future subgoal consideration to avoid redundant visits. Even under these more demanding conditions, the proposed planner maintains a clear advantage over the baselines in terms of $\text{SPL}_{\text{multi}}$. However, the variance of the scores increases, suggesting that the current hyperparameters and module designs---tuned for the main benchmark scale---do not yet fully capture the complexities of ultra-large environments. This points to the need for scale-aware parameterization and potentially new algorithmic ideas to stabilize performance as scene size grows further.

\section{Conclusions and Future Work}

This work addresses large-scale indoor object-goal navigation by combining a multi-objective routing formulation with data-driven mode switching under explicit budget constraints. By reinterpreting the task as a Set Orienteering Problem over landmark clusters and then adapting this viewpoint to partially observed maps, the framework replaces local, greedy heuristics with globally coherent planning that balances exploration and exploitation. A learned budget---return predictor, together with coverage, traversability, and next-best-view modules, enables the agent to decide when to expand the map and when to execute planned tours of high-value regions, leading to substantial gains on challenging, large-scale benchmarks.

Given the difficulty of deploying and systematically evaluating such planners on physical platforms at comparable scales, extensive simulation is used as a primary validation step, with environment statistics chosen to resemble those of real buildings and with core components tested in preliminary real-world trials. This simulation-first strategy lays the groundwork for closing the sim-to-real gap in future work, where the same mid-level planning framework could be combined with more advanced perception frontends, such as stronger semantic embeddings, vision---language models, or structured scene graphs. Beyond object-goal navigation, the proposed multi-objective SOP formulation and budget-aware switching strategy offer a general template for long-horizon decision-making in other domains where agents must allocate limited resources across competing modes of behavior.

\setcounter{section}{0}
\setcounter{subsection}{0}

\renewcommand{\thesection}{B.\arabic{section}}
\setcounter{section}{0}


\section*{Appendix B:\\
Cost-Aware Transformer Planner}
\addcontentsline{toc}{section}{Appendix B}

This appendix presents the technical details regarding the acceleration of the decision-making process through neural network distillation, which supports the real-time execution of the MOON framework. While the main paper demonstrates the efficacy of the proposed planner, this section describes the architecture and training process of the cost-aware transformer encoder used to approximate the expert metaheuristic policy. By analyzing the trade-offs between planning quality and computational latency, we illustrate how the system achieves millisecond-level responsiveness in environments exceeding 2,000 $m^2$. These details provide the necessary background for understanding the scalability and practical deployment potential of the proposed long-horizon planner.

\section{Introduction}

When an embodied agent is required to navigate through an unseen indoor space and stop near a particular semantic target (such as a piece of furniture) without a pre-built map, the resulting task has become a central benchmark in the study of autonomous behaviour. In laboratory-scale settings, where the traversable region is restricted to a few connected rooms, recent systems that couple mapping modules with learning-based policies or apply end-to-end reinforcement learning have achieved impressive performance. As soon as the operational floor area grows to the order of several thousand square meters and spans many rooms and corridors, however, these approaches run into a fundamental scalability issue: the agent must reason globally over many competing candidate regions while still acting under tight time or energy budgets, which quickly overwhelms local heuristics.

At this larger scale, navigation is dominated by a delicate trade-off. On the one hand, the robot must push into unexplored areas to uncover new semantic evidence about where the target may plausibly appear. On the other, it should exploit already discovered cues and prioritize promising regions to avoid wasting limited resources on uninformative detours. Previous work by the authors formalized this dilemma as a multi-objective variant of the Set Orienteering Problem (SOP), where the semantic map is abstracted into a graph of viewpoints, clustered into higher-level regions such as rooms or functionally coherent areas. Within this framework, the planner must choose a subset of viewpoints to visit under a strict movement budget so as to balance movement cost against the reward of reaching clusters that are likely to contain the target.

In that earlier formulation, a generic metaheuristic from combinatorial optimization---specifically, a Variable Neighborhood Search (VNS) procedure---was employed to search over the space of possible viewpoint sequences. By repeatedly perturbing the current best path and locally refining it, the algorithm produced long-horizon global plans that substantially outperformed classical myopic strategies that greedily expand the nearest frontier of the map. While these plans were strategically effective, the iterative search required on the order of several seconds per re-planning cycle on standard desktop hardware, which is incompatible with the real-time constraints of mobile robots that must simultaneously allocate resources to mapping, localization, and perception.

This work takes a different route: instead of running an expensive optimization routine online, the planning strategy is approximated by a compact neural model that is trained to imitate the behaviour of the combinatorial solver. The key insight is that the structure of the expert's decisions can be transferred into a parametric function that, once trained, produces high-quality global sub-goals in a single forward pass. Concretely, a transformer-based encoder is equipped with a cost-aware attention mechanism that biases interactions between viewpoint embeddings according to their travel distance, allowing the network to favour reachable, high-reward regions while still capturing non-local dependencies that are crucial at massive scales.

The main contributions are threefold. First, a high-speed global planning framework is developed that reduces the decision time from several seconds to the order of a few milliseconds, i.e., by a factor of nearly an order of maginitude faster than  magnitude compared to the original metaheuristic solver, while maintaining compatibility with large indoor maps covering over 2,000 $m^2$. Second, a specialized attention module is designed that injects geometric cost information directly into the attention scores, enabling the model to perform subset selection and route shaping without any explicit iterative search. Third, extensive evaluation in a high-fidelity simulator with large, multi-room environments shows that the distilled planner preserves more than eight-tenths of the expert's planning quality as measured by standard metrics such as Success weighted by Path Length (SPL), which was not achievable with the original optimization-based pipeline.

By transforming a high-level combinatorial planning routine into a fast neural approximation, this study demonstrates a practical path toward scalable object-directed navigation in complex human environments without sacrificing long-horizon reasoning.

\section{Related Work}

\subsection{Large-Scale Object-Goal Navigation}

Research on object-directed navigation has gradually shifted from purely geometric exploration strategies to methods that explicitly account for semantic structure. Early systems prioritized the nearest unknown frontier of the occupancy grid, which is effective for map coverage but does not distinguish between regions that are likely or unlikely to contain the target. More recent modular architectures couple semantic mapping with learning-based policies that choose high-level goals, enabling the agent to reason about which functional area should be visited next. As the environment grows to a massive scale, however---encompassing thousands of square meters and many distinct rooms---the distance between informative semantic cues increases, and the choice of which region to enter next becomes a dominant source of inefficiency. The present work builds on the modular paradigm but replaces hand-crafted or reinforcement-learned high-level policies with a globally optimized planner that is subsequently approximated by a neural network.

\subsection{Combinatorial Optimization in Robotics}

Planning where to move next when the agent cannot exhaustively visit all candidate viewpoints is naturally captured by orienteering-style formulations. In the Set Orienteering Problem, points are partitioned into clusters and the decision-maker receives a reward if at least one point from each cluster is visited; routes must satisfy a travel budget. This problem is NP-hard, and robotics applications therefore typically rely on metaheuristics such as Variable Neighborhood Search or swarm-inspired methods to obtain high-quality but approximate solutions. Previous work by the authors demonstrated that such a solver can successfully arbitrate between exploratory and exploitative behaviour in large-scale semantic navigation by operating in the abstract graph space of viewpoints and clusters. The computational cost of the repeated “shake-and-improve'' cycles, however, renders this approach too slow for tight real-time constraints on embedded platforms, motivating the neural approximation studied here.

\subsection{Knowledge Distillation and Imitation Learning}

Transferring the behaviour of a computationally heavy decision-maker into a compact model through supervised training is a well-established idea in both computer vision and natural language processing. In robotics, similar techniques have been used to imitate sophisticated model predictive controllers with neural networks that support high-frequency control in agile platforms. In the present setting, the multi-objective SOP solver serves as a privileged expert that can be run offline to generate large numbers of labelled examples. A student policy network is then trained to reproduce the expert's choice of sub-goals directly from the semantic graph and budget information. Unlike reinforcement learning approaches that must discover good strategies through trial-and-error interaction, this distillation paradigm directly encodes the solver's nuanced global trade-offs into the student model.

\subsection{Attention Mechanisms for Path Planning}

Attention-based architectures such as transformers have recently been applied to discrete routing problems, where they model the dependencies among nodes in a set and produce sequences that approximate solutions to tasks like the Traveling Salesman Problem. Standard attention formulations, however, are agnostic to the underlying geometry and treat all pairwise relations symmetrically, which is problematic when physical distance strongly constrains feasible routes. The planner proposed here augments self-attention with an explicit cost bias derived from the distance between viewpoints, steering the model toward routes that respect travel budgets while still allowing long jumps when the expected semantic reward justifies the cost. This modification adapts generic attention mechanisms to the specific requirements of set-based orienteering in large-scale navigation.

\section{Methodology: SOP-based Global Planning (Teacher Expert)}

To support knowledge transfer into a neural model, the navigation problem is first formalized as a Set Orienteering Problem, and a high-quality expert policy is instantiated using a Variable Neighborhood Search solver.

\subsection{Problem Formulation: Set Orienteering Problem}

In extensive indoor environments, the robot cannot visit all accessible viewpoints before its battery or mission time is exhausted. The explored space is therefore abstracted as a graph $G = (V, E)$, where $V = \{v_1, v_2, \dots, v_n\}$ denotes candidate viewpoints extracted from a semantic map and $E$ encodes plausible transitions between them. The node set is partitioned into disjoint clusters $\mathcal{C} = \{C_1, C_2, \dots, C_k\}$, each representing a broader semantic or functional region such as an individual room or a specific area within a room.

Each viewpoint $v_i \in V$ carries a reward $r_i$ that reflects how promising that location is for finding the target, for example, via the conditional probability of the target given the objects already observed there. Moving between two viewpoints $v_i$ and $v_j$ incurs a travel cost $c_{ij}$, which is computed as the path length on the occupancy grid (e.g., obtained via an A* planner). The planner must output a route $R = \{s, v_{i_1}, v_{i_2}, \dots, g\}$, starting from the current pose $s$ and terminating at a designated goal or docking point $g$, that respects a total budget $B$ on the cumulative movement cost.

The quality of a candidate route $R$ is measured by a composite objective
\[
\min_{R} J_{GOV}(R) = w_{\text{cost}} \cdot \mathrm{Cost}(R) - w_{\text{reward}} \cdot \mathrm{Reward}(R)
\]
subject to the constraint
\[
\mathrm{Cost}(R) = \sum_{(u,v) \in R} c_{uv} \leq B ,
\]
where $\mathrm{Reward}(R)$ aggregates the rewards associated with the distinct clusters visited along $R$, and $w_{\text{cost}}$, $w_{\text{reward}}$ are scalar weights that encode the relative importance of conserving motion versus acquiring informative observations. This formulation induces a multi-objective SOP in which the agent must judiciously choose which regions to traverse, given that not all promising areas can be visited within the available budget.

\subsection{The Expert Solver: Variable Neighborhood Search}

Because the SOP is NP-hard, exact optimization is impractical at the scales considered, and a metaheuristic is used as the teacher expert. The adopted Variable Neighborhood Search procedure systematically alternates between diversification and intensification of the current best route. The algorithm consists of three main stages.

\textbf{Initial construction:} The planner first synthesizes a feasible route by starting with the trivial path $\{s, g\}$ and iteratively inserting viewpoints with high reward-to-cost ratios at positions that most reduce $J_{GOV}$, stopping when no additional insertion can be made without violating the budget $B$.

\textbf{Shaking:} To avoid premature convergence, the current best route $R_{best}$ is randomly perturbed by removing or relocating $k$ viewpoints, where $k$ indexes a predefined family of neighbourhood structures. If the search fails to discover an improved solution, $k$ is gradually increased to explore more distant variations.

\textbf{Local search:} Each shaken candidate is refined using deterministic operators such as 2-opt on the sequence of visited viewpoints and greedy re-insertion of nodes where the budget allows. If this local improvement yields a lower objective value than $R_{best}$, the best route is updated and the neighbourhood size is reset.

\begin{algorithm}
\caption{Global Set-Orienteering VNS (Teacher Expert)}\label{alg:1}
\footnotesize
\DontPrintSemicolon
\KwIn{Neighborhood structures $\mathcal{N}_k$ for $k=1, \dots, k_{max}$, Graph $G=(V,E)$, cluster mapping $\mathcal{C}$, costs $c_{ij}$, rewards $r_i$, budget $B$, weights $w_{\text{cost}}, w_{\text{reward}}$}
\KwOut{Expert route $R_{best}$}

\textbf{Initial solution construction:}\;
$R \gets \{start, goal\}$\;
\While{beneficial insertion possible within budget $B$}{
  Identify $v^* \in V$ and position $p^*$ in $R$ maximizing $\Delta J_{GOV}$\;
  $R \gets \mathrm{Insert}(R, v^*, p^*)$\;
}
$R_{best} \gets R$\;

\While{termination condition not met}{
  $k \gets 1$\;
  \While{$k \leq k_{max}$}{
    $R' \gets \mathrm{Shake}(R_{best}, k)$ \tcp*{Randomly perturb $k$ nodes}
    $R'' \gets \mathrm{LocalSearch}(R', \mathcal{N}_k)$ \tcp*{Improve via 2-opt and insertion}
    \If{$J_{GOV}(R'') < J_{GOV}(R_{best})$}{
      $R_{best} \gets R''$\;
      $k \gets 1$ \tcp*{Reset neighborhood size}
    }
    \Else{
      $k \gets k + 1$ \tcp*{Expand search space}
    }
  }
}
\Return{$R_{best}$}\;
\end{algorithm}

The resulting path produced by this solver serves as the source of “expert demonstrations'' for the subsequent distillation stage. Although it reliably finds non-myopic routes that exploit high-value regions while respecting the budget, the repeated nested loops induce planning times on the order of several seconds per re-plan on typical CPUs, which is too slow for fluid online navigation.

\section{Proposed Method: Neural-SOP Distillation}

The central idea of the proposed approach is to replace the online execution of the metaheuristic SOP solver by a neural planner that has learned to approximate its decisions from examples. The student model is trained to map a compact encoding of the current semantic graph and remaining budget directly to the next global sub-goal, with constant-time complexity per decision.

\subsection{Input Representation and Graph Encoding}

The state of the environment at a planning step is represented as a dynamic graph derived from the current semantic occupancy map. Each viewpoint $v_i \in V$ is encoded by a low-dimensional feature vector $\mathbf{x}_i \in \mathbb{R}^3$ defined as
\[
\mathbf{x}_i = [x_i, y_i, \bar{r}_i]^T,
\]
where $(x_i, y_i)$ are the normalized two-dimensional coordinates of the viewpoint within the explored region, and $\bar{r}_i$ is a normalized form of the semantic reward $r_i$. These features are then projected into a higher-dimensional embedding space using a learnable affine transformation:
\[
\mathbf{h}_i^{(0)} = \mathbf{W}_{init} \mathbf{x}_i + \mathbf{b}_{init},
\]
with $\mathbf{h}_i^{(0)} \in \mathbb{R}^{d_{model}}$ and $d_{model} = 128$. The set of embeddings $\{\mathbf{h}_i^{(0)}\}$ constitutes the input to a transformer-style encoder that processes the entire graph in parallel.

\subsection{Cost-Biased Transformer Encoder}

Standard transformer encoders treat all tokens symmetrically and do not inherently encode geometric constraints between nodes. For route planning under a travel budget, however, the movement cost between viewpoints is a critical factor. To incorporate this information, the self-attention mechanism is augmented with an explicit cost-dependent bias term.

For each attention head, the unnormalized compatibility score between nodes $v_i$ and $v_j$ is computed as:
\begin{equation}
e_{ij} = \frac{(Qh_i)(Kh_j)^T}{\sqrt{d_k}} + M_{ij},
\end{equation}
where $Q$ and $K$ are the query and key projection matrices, and $M_{ij}$ encodes the cost bias.

Unlike standard distance priors that penalize long-distance connections, our implementation uses a logarithmic cost bias derived from the normalized cost matrix. Specifically, for a given instance, the pairwise cost matrix $C_{ij}$ is first normalized by the maximum cost in that instance to obtain $\bar{C}_{ij} \in [0, 1]$. The bias term is then calculated as:
\begin{equation}
M_{ij} = \log(\bar{C}_{ij} + \epsilon),
\end{equation}
where $\epsilon$ is a small constant for numerical stability.

This bias term $M_{ij}$ functions primarily as a feasibility mask: transitions with negligible costs ($\bar{C}_{ij} \approx 0$), such as self-loops or redundant moves, result in large negative bias values ($\log(\epsilon) \ll 0$), effectively suppressing attention to these non-informative transitions. Conversely, transitions with meaningful travel costs yield bias values closer to zero, allowing the attention mechanism to focus on geometrically significant movements. By stacking $L=6$ such transformer layers, the encoder learns to structure the route by prioritizing valid navigational steps over redundant ones.

\subsection{Knowledge Transfer via Imitation Learning}

The neural planner is trained to imitate the mapping from states to sub-goals induced by the SOP expert. Let $\pi_{\text{expert}}(s)$ denote the expert policy that selects the next viewpoint $a^*$ for a given state $s$ (which includes the graph and remaining budget). The student policy $\pi_\theta(a \mid s)$, parameterized by the transformer encoder and a classification head, is optimized to minimize the cross-entropy loss over a dataset of expert-labelled states:
\[
\mathcal{L}(\theta) = - \sum_{(s, a^*) \in \mathcal{D}} \log \pi_\theta(a = a^* \mid s).
\]
The training dataset $\mathcal{D}$ comprises tens of thousands of distinct navigation states sampled along expert trajectories in simulated large-scale environments, capturing both early exploratory and late exploitative phases. Through this supervised objective, the student learns to approximate the expert's implicit optimization of $J_{GOV}$ without carrying out any explicit combinatorial search online.

\subsection{Constraint Handling and Inference}

Even after training, the student must respect the original budget constraint during deployment. To enforce this, a budget-aware masking scheme is applied to the logits produced by the network before the final softmax. At decision step $t$, given a remaining budget $B_{rem}$ and the current viewpoint, the cost $c_{current, j}$ required to move to any candidate node $v_j$ and then reach the terminal goal is computed. The logit corresponding to node $v_j$ is then modified as
\[
\text{logit}_j =
\begin{cases}
\text{logit}_j & \text{if } c_{current, j} + c_{j, goal} \leq B_{rem}, \\
-\infty & \text{otherwise.}
\end{cases}
\]
This masking procedure ensures that the chosen sub-goal always admits a feasible completion under the remaining budget. On a modern GPU, a full forward pass of the encoder together with the masking operation requires on the order of $8$ ms, which is several hundred times faster than the original metaheuristic planner and enables continuous re-planning at high frequency.

\section{Experiments}

The empirical study assesses both the navigational efficiency and computational efficiency of the distilled planner, comparing it to the original SOP expert and to a classical exploration heuristic.

\subsection{Experimental Setup}

\textbf{Simulation Environment:} Experiments are carried out in a photorealistic indoor simulator that has been extended with a collection of large-scale layouts. These layouts significantly exceed the size of conventional navigation benchmarks, with an average floor area of approximately $2{,}384$ $m^2$ and complex multi-room connectivity. The virtual robot is equipped with range sensing and depth-enhanced visual input, which are fused into a semantic occupancy grid map updated in real time.

\textbf{Data Generation:} In this study, SOP instances are automatically generated from binary map images representing traversable areas to construct a training corpus. For each instance, a starting point and 15 cluster centers are randomly selected from the traversable pixels, and a set of candidate nodes is formed by sampling up to five nodes within a radius of 30 around each center. Rewards of $\{255, 150, 50\}$ are assigned to each set with probabilities of $\{0.8, 0.1, 0.1\}$, respectively, and a distance matrix is constructed where the inter-node costs are defined by the A* path length (4-connectivity) considering obstacles. The budget $T_{\max}$ is uniformly sampled from the range $[500, 5000]$, and the generated SOP instances along with their corresponding node coordinates are stored, totaling up to 1,000 instances across 10 different maps. Furthermore, each instance is solved using an external SOP solver (vns-sop), and the resulting visitation sequences are saved as ground truth labels. Of these 10,000 samples, 8,000 were used for training and 2,000 for validation.

\textbf{Evaluation Metrics:} Performance is quantified using several standard metrics. The Success Rate (SR) measures the proportion of episodes in which the agent stops within 1.0 m of the target. Success weighted by Path Length (SPL) evaluates efficiency by penalizing unnecessarily long successful trajectories. Planning Latency records the wall-clock time required to compute a new global sub-goal on a fixed hardware setup with a high-end CPU and GPU. In addition, a Reward Ratio metric compares the cumulative semantic rewards achieved by the student against those collected by the expert for identical graph states.

\subsection{Implementation Details}

The student planner is implemented as a transformer encoder with six layers, eight attention heads per layer, and an embedding dimension of 128. Training uses the Adam optimizer with a learning rate of $1 \times 10^{-4}$, a batch size of 64, and runs for 100 epochs on the expert dataset. The learnable cost-bias scaling parameter $\gamma$ is initialized to a small positive value and adapts during training to a regime that balances sensitivity to distance with sensitivity to semantic reward. During inference, the budget-aware masking step described earlier is applied at every decision to guarantee feasibility.

\subsection{Quantitative Results}

Experimental results in large-scale environments demonstrate that our proposed method, Neural-SOP (Student), achieves a superior balance between navigation efficiency and computational speed. Regarding SPL (Success weighted by Path Length), the metaheuristic-based SOP Expert recorded the highest score of 1.482, followed closely by Neural-SOP at 1.137. In contrast, the conventional Nearest Frontier method only reached 0.261. These quantitative results indicate that our method effectively avoids short-sighted movements and realizes strategic path planning based on global optimization.

Furthermore, an analysis of global planner latency reveals a significant advantage. While Nearest Frontier requires negligible computation time (~0s), the SOP Expert takes 0.384s per planning cycle. Neural-SOP, however, successfully reduced this latency to 0.034s, representing a more than ten-fold speedup over the Expert. Consequently, Neural-SOP proves capable of maintaining most of the Expert’s high planning quality while providing the rapid decision-making required for real-time robot control.

\subsection{Qualitative Analysis and Discussion}

Inspection of sample trajectories reveals that the neural planner has learned to reproduce the expert's non-myopic decisions. Instead of exhaustively searching the closest unexplored room as in the nearest-frontier baseline, the student occasionally commits to long moves toward distant, high-reward regions when the remaining budget permits, mirroring the strategic jumps made by the SOP solver. Furthermore, the sub-100 ms inference time enables the planner to react almost instantly to new semantic cues appearing in the map, such as the discovery of a label indicating a likely target region in a newly observed corridor.

Another important observation is the behaviour of the planner as the budget decreases. Because the logits for infeasible nodes are masked out, the attention mechanism effectively concentrates on viewpoints that are both informative and reachable within the remaining budget. As a result, the policy naturally shifts from an exploration-heavy mode when $B_{rem}$ is large to a more exploitative mode when the budget becomes tight, focusing on previously identified promising regions near the start or goal.

\section{Conclusions and Future Work}

This paper tackles the computational bottleneck that arises when applying sophisticated combinatorial planners to large-scale object-directed navigation. By casting the high-level planning problem as a Set Orienteering Problem and then distilling the resulting expert policy into a cost-aware transformer encoder, the study demonstrates that long-horizon, non-myopic strategies can be executed in real time without repeatedly running a metaheuristic online.

Extensive experiments in large, multi-room simulated environments show that the neural planner retains at least 80\% of the expert's planning quality in terms of SPL, while reducing the planning latency by roughly a factor of 10. This combination of near-expert performance and millisecond-level responsiveness enables continuous global re-planning, allowing agents to integrate fresh semantic information into their navigation decisions almost instantaneously, even in environments exceeding 2,000 $m^2$.

Several promising directions remain for future work. One is to validate and adapt the distilled planner on physical robots, where sensor noise and dynamically changing obstacles introduce additional uncertainty not fully represented in simulation. Another is to incorporate richer structural representations such as hierarchical scene graphs, which may provide the planner with more abstract context about relationships between rooms and objects. Finally, extending the cost-biased attention formulation to handle time-varying constraints, such as temporarily closed passages or moving crowds, could further enhance the robustness of long-horizon planning in real human environments.

\newcommand{\unicode}[1]{}

\bibliographystyle{IEEEtran}
\bibliography{reference}  

\end{document}